\documentclass[]{llncs}
\usepackage{graphicx}
\usepackage{adjustbox}
\usepackage{bbm}
\usepackage{color,soul}

\begin{document}
\title{Evaluating deep tracking models for player tracking in broadcast ice hockey video }
%
%
\author{Kanav Vats \and
Mehrnaz Fani \and
David A. Clausi \and
John S. Zelek}
%
%
\institute{Department of Systems Design Engineering, University of Waterloo \\ \email{\{k2vats, mfani, dclausi, jzelek\}@uwaterloo.ca}}
\maketitle              
\begin{abstract}

Tracking and identifying players is an important problem in computer vision based ice hockey analytics. Player tracking is a challenging problem since the motion of players in hockey is fast-paced and non-linear. There is also significant player-player and player-board occlusion, camera panning and zooming in hockey broadcast video. Prior published research perform player tracking with the help of handcrafted features for player detection and re-identification. Although commercial solutions  for hockey player tracking exist, to the best of our knowledge, no network architectures used, training data or performance metrics are publicly reported. There is currently no published work for hockey player tracking making use of the recent advancements in deep learning while also reporting the current accuracy metrics used in literature. Therefore, in this paper we compare and contrast several state-of-the-art tracking algorithms and analyze their performance and failure modes in ice hockey.

\end{abstract}
\section{Introduction}
Ice hockey is played by an estimated $1.8$ million people worldwide~\cite{iihf2018survey}. As a team sport, the positioning of the players and puck on the ice are critical to team offensive and defensive strategy~\cite{thomas2006impact}. The location of players on the ice  is essential for hockey analysts for determining the location of play and analyzing game strategy and events.
In ice hockey, prior published research \cite{okuma,cai_hockey} perform player tracking with the help of handcrafted features for player detection and re-identification. Okuma \textit{et al.} \cite{okuma} track hockey players by introducing a particle filter combined with mixture particle filter (MPF) framework \cite{mixture_tracking}, along with an Adaboost \cite{adaboost} player detector. The MPF framework \cite{mixture_tracking} allows the particle filter framework to handle multi-modality by modelling the posterior state distributions of $M$ objects as an $M$ component mixture. A disadvantage of the MPF framework is that the particles merge and split in the process and leads to loss of identities. Moreover, the algorithm does not have any mechanism to prevent identity switches and lost identities of players after occlusions. Cai \textit{et al.} \cite{cai_hockey} improve upon \cite{okuma} by using a bipartite matching for associating observations with targets instead of using the mixture particle filter framework. However, the algorithm is not trained or tested on broadcast videos, but performs tracking in the rink coordinate system after a manual homography calculation. \par
Remarking that there is a lack of publicly available research for tracking ice hockey players making use of recent advancements in deep learning, in this paper we track and identify hockey players in broadcast NHL videos and analyze performance of several state-of-the-art deep tracking models on the ice hockey dataset. We also annotate and introduce a new hockey player tracking dataset on which the deep tracking models are tested.

\section{Related work}
There are a number of recent studies dealing with player tracking in basketball \cite{Sangesa2019SingleCameraBT,lu2013,ZHANG2020107260} and soccer \cite{Theagarajan_2021,hurault,Theiner_2022_WACV,Gadde_2022_WACV}. For basketball player tracking, Sangüesa \textit{et al.} \cite{Sangesa2019SingleCameraBT} demonstrated that deep features perform better than classical handcrafted features for basketball player tracking. Lu \textit{et al.} \cite{lu2013} perform player tracking in basketball using a Kalman filter by making the assumption that the relationship between time and player’s locations is approximately linear in a short time interval. Zhang \textit{et al.} \cite{ZHANG2020107260} perform basketball player tracking in a multi camera setting. \par 

In soccer, Theagarajan \textit{et al.} \cite{Theagarajan_2021} track players  using the deep SORT algorithm \cite{Wojke2017simple} for generating tactical analysis and ball possession statistics . Hurault \textit{et al.}  \cite{hurault} introduce a self-supervised detection algorithm to detect small soccer players and track players in non-broadcast settings using a triplet loss trained re-identification mechanism, with embeddings obtained from the detector itself. Theiner \textit{et al.} \cite{Theiner_2022_WACV} present a pipeline to extract player position data on the soccer field from video. The player tracking was performed with the help of CenterTrack \cite{zhou2020tracking}. However, the major focus of the work was on detection accuracy rather than tracking and identification. Gadde \textit{et al.} \cite{Gadde_2022_WACV} use a weakly supervised transductive approach for player detection in soccer broadcast videos by treating player detection as a domain adaptation problem. The dataset used is generated with the help of the deep SORT algorithm \cite{Wojke2017simple}.

\section{Methodology}
We experimented with five state-of-the-art tracking algorithms \cite{Bewley2016_sort,Wojke2017simple,zhang2020fair,tracktor_2019_ICCV,Braso_2020_CVPR} on the hockey player tracking dataset. The algorithms include four online tracking algorithms \cite{Bewley2016_sort,Wojke2017simple,zhang2020fair,tracktor_2019_ICCV} and one offline tracking algorithm \cite{Braso_2020_CVPR}.  SORT \cite{Bewley2016_sort}, deep SORT \cite{Wojke2017simple} and MOT Neural Solver \cite{Braso_2020_CVPR} are tracking by detection (TBD) algorithms. Tracktor \cite{tracktor_2019_ICCV} and FairMOT \cite{zhang2020fair} are joint detection and tracking (JDT) algorithms. \par

Tracking by detection (TBD) is a widely used approach for multi-object tracking. TBD consists of three steps: (1) detecting objects (hockey players in our case) frame-by-frame in the video (2) calculating affinity between detected objects (3) inference - linking player detections using calculated affinities to produce tracks. Concretely, in TBD, the input is a set of object detections $O = \{o_1,.....o_n\}$, where $n$ denotes the total number of detections in all video frames. A detection $o_i$ is represented by $\{x_i,y_i,w_i,h_i, I_i,t_i\}$, where $x_i,y_i,w_i,h_i$ denotes the coordinates, width, and height of the detection bounding box. $I_i$ and $t_i$ represent the image pixels and timestamp corresponding to the detection. Affinity calculation consists of calculating affinity between detections $o_i$ by obtaining appropriate features. The features can be simple intersection over union (IOU) based \cite{Bewley2016_sort} or using deep networks \cite{Wojke2018deep}. After affinity calculation, a set of trajectories $T = \{T_1,T_2...T_m\}$  is found that best explains $O$ where each $T_i$ is a time-ordered set of observations. This is done through an appropriate inference technique. Two widely used inference techniques are filtering \cite{Bewley2016_sort,Wojke2018deep} and graphical formulation \cite{Braso_2020_CVPR}. As an example of graphical formulation, the MOT Neural Solver \cite{Braso_2020_CVPR} models the tracking problem as an undirected graph $G=(V,E)$ , where $V=\{1,2, ..., n\}$ is the set of $n$ nodes for $n$ player detections for all video frames. In the edge set $E$, every pair of detections is connected so that trajectories with missed detections can be recovered. The problem of tracking is posed as splitting the graph into disconnected components where each component is a trajectory $T_i$. After computing each node embedding and edge embedding using a CNN (affinity calculation), the model then solves a graph message passing problem. The message passing algorithm classifies whether an edge between two nodes in the graph belongs to the same player trajectory. \par

\begin{table*}[!t]

    \centering
    \caption[Tracking algorithms compared for hockey player tracking.]{Tracking algorithms compared for hockey player tracking. }
    \footnotesize
       \begin{adjustbox}{width=1\textwidth}
    \setlength{\tabcolsep}{0.2cm}
    \begin{tabular}{c|c}\hline
  
       Algorithm & Description \\\hline
       SORT \cite{Bewley2016_sort} & Kalman filter with simple IOU based re-id. \\
      Deep SORT \cite{Wojke2017simple} &  Kalman filter with deep CNN based re-id. \\
       Tracktor \cite{tracktor_2019_ICCV} & JDT algorithm with separate detection and re-id networks. \\
        FairMOT \cite{zhang2020fair} & JDT algorithm with combined object detection and re-id network.     \\
        MOT Neural Solver \cite{Braso_2020_CVPR}  & Tracking using graph message passing with edge classification.
    \end{tabular}
      \end{adjustbox}
    \label{table:compared_algorithms}
\end{table*}

 Joint detection and tracking (JDT) \cite{tracktor_2019_ICCV,zhang2020fair} is the latest trend in multi-object tracking research. These methods either (1) Convert an object detector to a tracker by estimating the location of a bounding box in the adjacent frames \cite{tracktor_2019_ICCV} or (2) Perform detection and re-identification using a single network \cite{zhang2020fair}. Bergmann \textit{et al.} \cite{tracktor_2019_ICCV} use the bounding box regressor of a Faster RCNN \cite{fasterrcnn} detector to regress the position of a person in the next frame. The re-identification is performed using a separate siamese network.  Zhang \textit{et al.} \cite{zhang2020fair} perform object detection and re-identification with the same network using separate detection and re-identification branches. The differences and similarities between the five tracking algorithms are summarized in Table \ref{table:compared_algorithms}. We refer the readers to the publications of the respective tracking papers \cite{Bewley2016_sort,Wojke2017simple,zhang2020fair,tracktor_2019_ICCV,Braso_2020_CVPR} for more detail.

 \begin{figure}[t]
\begin{center}
\includegraphics[width=\linewidth]{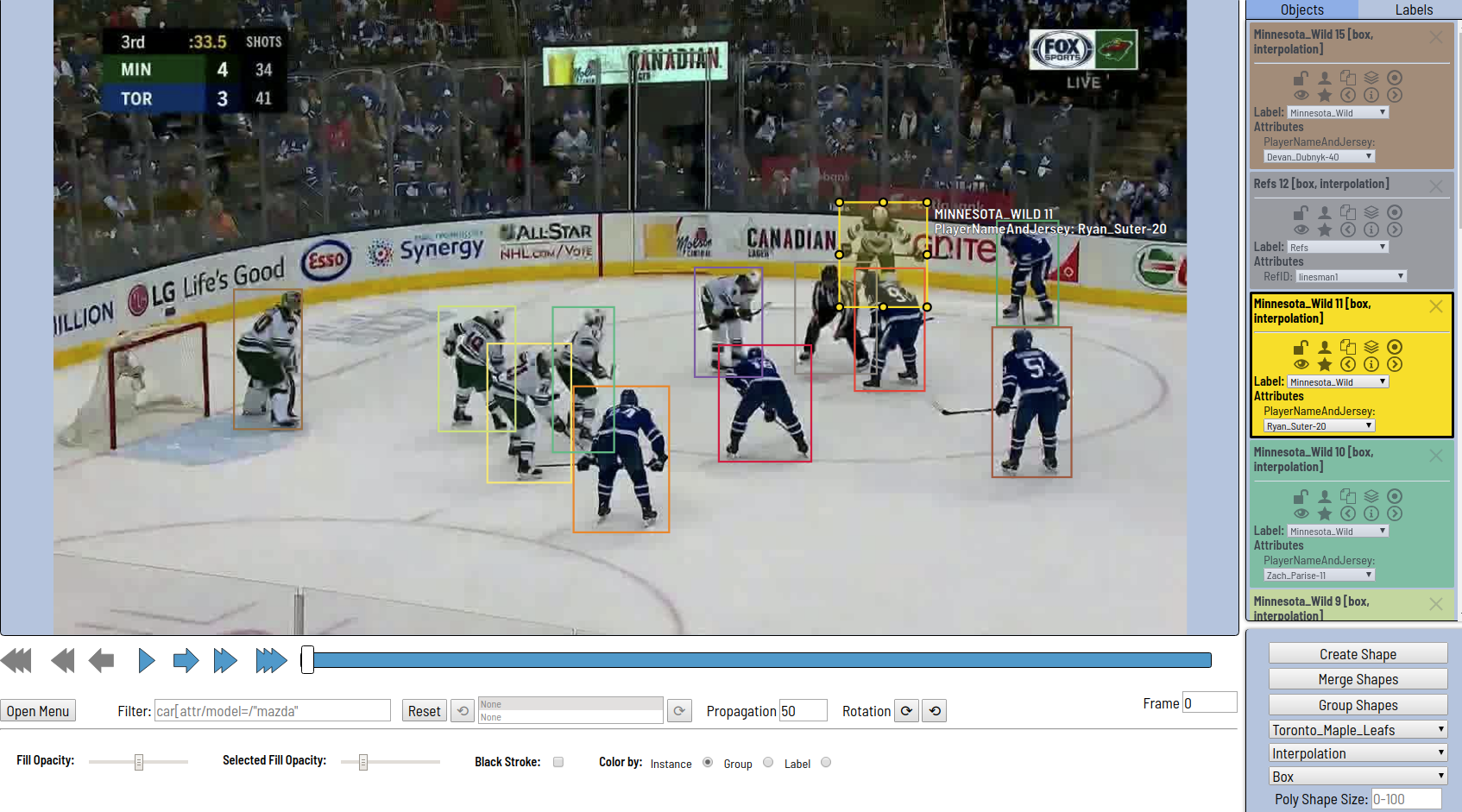}
\end{center}
  \caption[CVAT tool used for tracking annotations]{CVAT tool used for tracking annotations. The tool offers the ability to annotate bounding boxes with each box having one label - home or away team. Each player bounding box has player name and jersey number as attributes. CVAT also offers an interpolation mode which alleviates the need to draw bounding boxes multiple times for adjacent frames. }
\label{fig:cvat}
\end{figure}

 \section{Dataset}
\begin{figure}[t]
\begin{center}
\includegraphics[width=\linewidth]{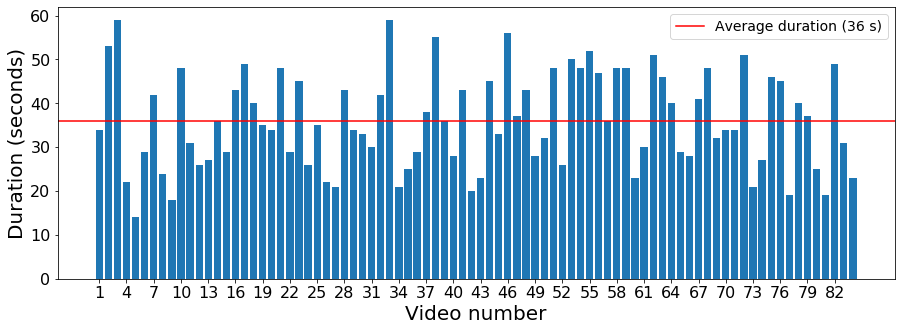}
\end{center}
  \caption[Duration of videos in the player tracking dataset.]{Duration of videos in the player tracking dataset. The average clip duration is $36$ seconds. The red horizontal line represents the average clip duration. }
\label{fig:clip_lengths}
\end{figure}

The player tracking dataset consists of a total of 84 broadcast NHL game clips with a frame rate of 30 frames per second (fps) and resolution of $1280 \times 720$ pixels. The average clip duration is $36$ seconds. The 84 video clips in the dataset are extracted from 25 NHL games. The duration of the clips is shown in Fig. \ref{fig:clip_lengths}. Each video frame in a clip is annotated with player and referee bounding boxes and player identity consisting of player name and jersey number. The annotation is carried out with the help of the open source computer vision annotation tool (CVAT) \footnote{Found online at: \url{https://github.com/openvinotoolkit/cvat}}. An illustration of an annotation job using the CVAT tool is shown in Fig. \ref{fig:cvat}.   The dataset is split such that 58 clips are used for training, 13 clips for validation, and 13 clips for testing. To prevent any game-level bias affecting the results, the split is made at the game level, such that the training clips are obtained from 17 games, validation clips from 4 games and test split from 4 games respectively. \\
 Table \ref{table:size_comparison} compares the size of the dataset with other tracking datasets in literature. The hockey player tracking dataset is comparable in size with other tracking datasets used in literature. As compared to pedestrian datsets (MOT 16 \cite{Milan2016MOT16AB} and MOT20 \cite{Dendorfer2020MOT20AB}), the bounding boxes per frame is less in our dataset since the maximum number of players on the screen can be $12$, with usually less than $12$ players actually in broadcast camera field of view (FOV). The NHL game videos used to create this dataset have been obtained from  Stathletes Inc. with permission.

\subsection{Annotation process}

 15 annotators annotated the whole dataset using the CVAT tool. The average time taken to annotate one minute of video is $10.45$ minutes. The total time taken to annotate all 84 videos is $527$ minutes. The manual annotation was done such that a bounding box as tight as possible was drawn around a player/referee. Linear interpolation was used to interpolate bounding box positions. Additionally, unlike other tracking datasets such as MOT16 \cite{Milan2016MOT16AB} and MOT20  \cite{Dendorfer2020MOT20AB}, the same ground truth identity was assigned to a player leaving camera FOV at a particular frame and re-entering after some time. If a player was occluded by board or another player, the bounding box was annotated based on the best guess of the tightest box enclosing the full body of the player. For quality control, all bounding boxes were checked  to make sure each box has label-name(name of the player ). When a player enters/exits the scene, his bounding box was labeled even if he was partially in camera FOV. Whenever players were occluded by other players, revision of annotations was performed to ensure high quality.

\begin{table*}[!t]

    \centering
    \caption[Comparison of hockey tracking dataset with other tracking datasets in the literature.]{Comparison of hockey tracking dataset with other tracking datasets in literature.  Our hockey player tracking dataset is comparable to other multi-object tracking datasets commonly used in literature. }
    \footnotesize
      \begin{adjustbox}{width=1\textwidth}
    \setlength{\tabcolsep}{0.2cm}
    \begin{tabular}{c|c|c|c|c}\hline
  
       Dataset & Videos/sequences & Frames& Bounding boxes & Domain \\\hline
     
     MOT16 \cite{Milan2016MOT16AB}  & $14$ &  $11,235$ & $292,733$ & Pedestrians \\
      MOT20 \cite{Dendorfer2020MOT20AB} & $8$ &  $13,410$ & $2,102,385$ & Crowded pedestrian scenes \\
       KITTI-T \cite{kitti}  & $50$ &  $10,870$ & $65,213$ & Autonomous driving \\
        Ours  & $84$ &  $91,807$ & $773,545$ & Ice hockey players \\

    \end{tabular}
     \end{adjustbox}
    \label{table:size_comparison}
\end{table*}

\section{Results}
 Player detection is performed using a Faster-RCNN network \cite{NIPS2015_14bfa6bb} with a ResNet50 based Feature Pyramid Network (FPN) backbone \cite{fpn} pre-trained on the COCO dataset - a large scale object detection, segmentation, and captioning dataset, popular in computer vision \cite{coco} and fine tuned on the hockey tracking dataset. The object detector obtains an average precision (AP) of $70.2$ on the test videos (Table \ref{table:player_det_results}). The accuracy metrics for tracking used are the CLEAR MOT metrics \cite{mot} and Identification F1 score (IDF1) \cite{idf1}. A ground truth object missed by the trackers is called a false negative (FN) whereas a false alarm is called a false positive (FP). For any tracker, a low number of false positives (FP) and false negatives (FN) are favoured. An important metric is the number of identity switches (IDSW), which occurs when a ground truth ID $i$ is assigned a tracked ID $j$ when the last known assignment ID was $k \ne j$. A low number of identity switches is an indicator of accurate tracking performance. For sports player tracking, the IDF1 is considered a better accuracy measure than Multi Object Tracking accuracy (MOTA) since it measures how consistently the identity of a tracked object is preserved with respect to the ground truth identity. The overall  results are shown in Table \ref{table:tracking_results}. The best tracking performance  is achieved using the MOT Neural Solver tracking model \cite{Braso_2020_CVPR} re-trained on the hockey dataset. The MOT Neural Solver model obtains the highest MOTA score of $94.5$ and IDF1 score of $62.9$ on the test videos.

\begin{table*}[!t]

    \centering
    \caption[Comparison of the overall tracking performance on test videos.]{Comparison of the overall tracking performance on test videos of the hockey player tracking dataset. ($\downarrow$ means lower is better, $\uparrow$ mean higher is better) }
    \footnotesize
      \begin{adjustbox}{width=1\textwidth}
    \setlength{\tabcolsep}{0.2cm}
    \begin{tabular}{c|c|c|c|c|c}\hline
  
       Method & IDF1$\uparrow$ & MOTA $\uparrow$& ID-switches $\downarrow$ & False positives (FP)$\downarrow$ & False negatives (FN) $\downarrow$\\\hline
     
     SORT \cite{Bewley2016_sort} & $53.7$ &  $92.4$ & $673$ & $2403$ & $5826$\\
      Deep SORT \cite{Wojke2017simple} & $59.3$ &  $94.2$ & $528$ & $1881$ & $4334$\\
       Tracktor \cite{tracktor_2019_ICCV} & $56.5$ &  $94.4$ & $687$ & $1706$ & $\textbf{4216}$\\
        FairMOT \cite{zhang2020fair} & $61.5$ &  $91.9$ & $768$ & $\textbf{1179}$ & $7568$\\
        MOT Neural Solver \cite{Braso_2020_CVPR}  & $\textbf{62.9}$ &  $\textbf{94.5}$ & $\textbf{431}$ & $1653$ & $4394$\\
   
    \end{tabular}
     \end{adjustbox}
    \label{table:tracking_results}
\end{table*}

\begin{figure}[t]
\begin{center}
\includegraphics[width=\linewidth]{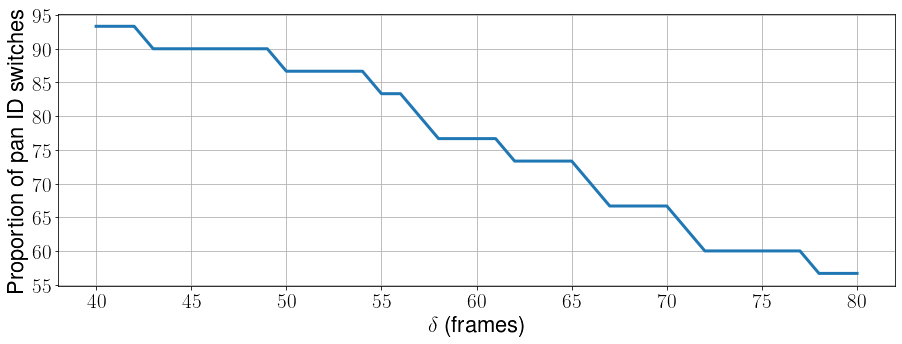}
\end{center}
  \caption[Proportion of pan identity switches vs. $\delta$ plot for video number $9$.]{Proportion of pan identity switches vs. $\delta$ plot for video number $9$. Majority of the identity switches ($~90\%$ at a threshold of $\delta= 40$ frames) occur due to camera panning, which is the main cause of error.}
\label{fig:vid_9_pidsw}
\end{figure}

\begin{table}[!t]

    \centering
    \caption[Player detection results on the test videos.]{Player detection results on the test videos. $AP$ stands for Average Precision. $AP_{50}$ and $AP_{75}$ are the average precision at an IOU of 0.5 and 0.75 respectively.}
    \footnotesize
     \begin{adjustbox}{width=0.3\textwidth}
    \setlength{\tabcolsep}{0.2cm}
    \begin{tabular}{c|c|c}\hline
  
         $AP$ & $AP_{50}$ & $AP_{75}$   \\\hline
     
    $70.2$ &  $95.9$    & $87.5$ 

    \end{tabular}
    \end{adjustbox}
    \label{table:player_det_results}
\end{table} 
\begin{figure}[t]
\begin{center}
\includegraphics[width=\linewidth]{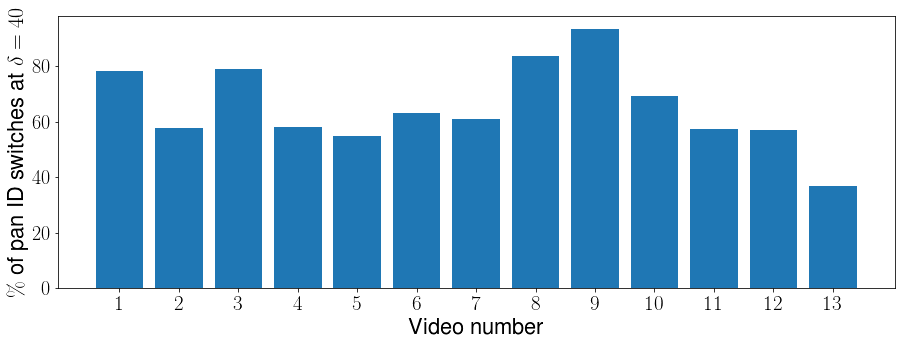}
\end{center}
  \caption[Proportion of pan-identity switches at a threshold of $\delta = 40$ frames.]{Proportion of pan-identity switches for all videos at a threshold of $\delta = 40$ frames. On average, pan-identity switches account for $65\%$ of identity switches. }
\label{fig:all_pidsw}
\end{figure}

\section{Discussion}
From Table \ref{table:tracking_results} it can be seen that the MOTA score of all methods is above $90\%$. This is because MOTA is calculated as
\begin{equation}
\label{MOT accuracy equation}
    MOTA = 1 - \frac{\Sigma_{t}(FN_{t} + FP_{t} + IDSW_{t})}{\Sigma_t GT_{t}}
\end{equation} 
where $t$ is the frame index and $GT$ is the number of ground truth objects. MOTA metric counts detection errors through the sum $FP+FN$ and association errors through $IDSWs$. Since false positives (FP) and false negatives (FN) heavily rely on the performance of the player detector, the MOTA metric highly depends on the performance of the detector. For hockey player tracking, the player detection accuracy is high  because of the sufficiently large size of players in broadcast video and a reasonable number of players and referees (with a fixed upper limit) to detect in the frame. Therefore, the MOTA score for all methods is high.\par
The SORT \cite{Bewley2016_sort} algorithm obtains the least IDF1 score and the highest number of identity switches. This is due to the linear motion model assumption and simple IOU score for re-identification. Deep SORT \cite{Wojke2018deep}, on the other hand uses features obtained from deep network for re-identification resulting in better IDF1 score and lower identity switches. For JDT based networks, performing detection and re-identification with a single network using a multi-task loss performs better than having separate networks for detection and re-id tasks, evident by better performance of FairMOT \cite{zhang2020fair} compared to Tracktor \cite{tracktor_2019_ICCV}.  JDT tracking algorithms, however, \cite{zhang2020fair,tracktor_2019_ICCV} do not not show any significant improvement over deep SORT evident by lower identity switches of deep SORT in comparison. The MOT Neural Solver method achieves the highest IDF1 score of $62.9$ and significantly lower identity switches than the other methods. This is because the other trackers use a linear motion model assumption which does not perform well with the motion of hockey players. Sharp changes in player motion often leads to identity switches. The MOT Neural Solver model, in contrast, has no such assumptions since it poses tracking as a graph edge classification problem.  \par


\begin{table*}[!t]

    \caption[Tracking performance of MOT Neural Solver model for the 13 test videos]{Tracking performance of MOT Neural Solver model for the 13 test videos ($\downarrow$ means lower is better, $\uparrow$ means higher is better). }
    \footnotesize
  \begin{adjustbox}{width=1\textwidth}
    \setlength{\tabcolsep}{0.2cm}
    \begin{tabular}{c|c|c|c|c|c|c}\hline
  
       Video \# & IDF1$\uparrow$ & MOTA $\uparrow$& ID-switches $\downarrow$ & False positives (FP)$\downarrow$ & False negatives (FN) $\downarrow$ & Duration (sec.)\\\hline
     
     1  & $78.53$ &  $94.95$ & $23$ & $100$ & $269$ & $36$\\
      2  & $61.49$ &  $93.29$ & $26$ & $48$ & $519$& $29$\\
       3  & $55.83$ &  $95.85$ & $43$ & $197$ & $189$ & $43$\\
        4  & $67.22$ &  $95.50$ & $31$ & $77$ & $501$ & $49$\\
    5   & $72.60$ &  $91.42$ & $40$ & $222$ & $510$ & $40$\\
      6  & $66.66$ &  $90.93$ & $38$ & $301$ & $419$ & $35$\\
      7  & $49.02$ &  $94.89$ & $59$ & $125$ & $465$ & $48$\\
       8  & $50.06$ &  $92.02$ & $31$ & $267$ & $220$ & $34$\\
        9  & $53.33$ &  $96.67$ & $30$ & $48$ & $128$ & $29$\\
    10   & $55.91$ &  $95.30$ & $26$ & $65$ & $193$ & $26$\\
      11  & $56.52$ &  $96.03$ & $40$ & $31$ & $477$ & $45$\\
        12  & $87.41$ &  $94.98$ & $14$ & $141$ & $252$ & $35$\\
    13   & $62.98$ &  $94.77$ & $30$ & $31$ & $252$ & $22$\\
   
    \end{tabular}
         \end{adjustbox}
    \label{table:video_wise_track}
\end{table*}

Table \ref{table:video_wise_track} shows the performance of the MOT Neural solver for each of the 13 test videos. We do a failure analysis to determine the cause of identity switches and low IDF1 score in some videos. The major sources of identity switches are severe occlusions and players going out of the camera FOV (due to camera panning and/or player movement). We define a pan-identity switch as an identity switch resulting from a player leaving and re-entering camera FOV due to camera panning. It is very difficult for the tracking model to maintain identity in these situations since players of the same team look identical with features such as, jersey color, helmet model, visor model, stick model, glove model, skate model, tape color etc unidentifiable from bounding boxes cropped from 720p broadcast clips. During a pan-identity switch, a player going out of the camera FOV at a particular point in screen coordinates can re-enter at any other point. We estimate the proportion of pan-identity switches to determine the contribution of panning to total identity switches. \par
To estimate the number of pan-identity switches, since we have quality annotations, we make the assumption that the ground truth annotations are accurate and there are no missing annotations in the ground truth. Based on this assumption, there is a significant time gap between two consecutive annotated detections of a player only when the player leaves the camera FOV and comes back again. Let  $T_{gt} = \{o_1, o_2,..., o_n \}$ represent a ground truth tracklet, where  $o_i = \{x_i,y_i,w_i,h_t,I_i, t_i\}$ represents a ground truth detection. A pan-identity switch is expected to occur during tracking when the difference between timestamps (in frames) of two consecutive ground truth detections $i$ and $j$ is greater than a sufficiently large threshold $\delta$.  That is \begin{equation}
    (t_i - t_j) > \delta 
\end{equation}
Therefore, the total number of pan-identity switches in a video is approximately calculated as
\begin{equation}
    \sum_G \mathbbm{1}( t_i - t_j > \delta ) 
\end{equation}
where the summation is carried out over all ground truth trajectories and $\mathbbm{1}$ is an indicator function. Consider the video number $9$ in Table \ref{table:video_wise_track} having $30$ identity switches and a low IDF1 of $53.33$. We plot the proportion of pan identity switches, that is \begin{equation}
     = \frac{\sum_G \mathbbm{1}( t_i - t_j > \delta )}{IDSWs} 
\end{equation}
against $\delta$, where $\delta$ varies between $40$ and $80$ frames.  From Fig. \ref{fig:vid_9_pidsw} it can be seen that majority of the identity switches ($~90\%$ at a threshold of $\delta= 40$ frames) occur due to camera panning. Visually investigating the video confirmed the statement. Fig. \ref{fig:all_pidsw} shows the proportion of pan-identity switches for all videos at a threshold of $\delta = 40$ frames. On average, pan identity switches account for $65\%$ of identity switches in the videos. This shows that the tracking model is able to tackle a majority of other sources of errors which include minor occlusions and lack of detections. The primary source or errors are pan-identity switches and extremely cluttered scenes.

\section{Conclusion}

In this paper, we test five state-of-the-art tracking algorithms on the ice hockey dataset and analyzed their performance. From the performance of trackers we infer that trackers with a linear motion model do not perform well on hockey dataset, evident by the high number of identity switches occurring in models with linear motion assumption. The best performance is obtained by the MOT neural solver model \cite{Braso_2020_CVPR}, that uses a graph based approach towards tracking without any linear motion model assumption. Also, the IDF1 metric is a better metric for hockey player tracking since the MOTA metric is heavily influenced by player detection accuracy. We find that the main source of error in hockey player tracking in broadcast video are pan-identity switches - identity switches results due to players going outside the broadcast camera FOV.

\section{Acknowledgments}

This work was supported by Stathletes through the Mitacs Accelerate Program and the Natural Sciences
and Engineering Research Council of Canada (NSERC). We also acknowledge Compute
Canada for hardware support. 

\bibliographystyle{splncs04}
\bibliography{bibliography}
%




\end{document}